\pdfoutput=1
\documentclass[11pt]{article}

\usepackage{EACL2023}

\usepackage{times}
\usepackage{latexsym}
\usepackage{graphicx}

\usepackage[T1]{fontenc}
\usepackage[utf8]{inputenc}

\usepackage{microtype}

\usepackage{inconsolata}

\usepackage{multirow, multicol}
\usepackage{amsmath}
\usepackage[inline]{enumitem}
\usepackage{csquotes}
\usepackage{tipa}

\title{Detecting Lexical Borrowings from Dominant Languages\\ in Multilingual Wordlists}
\author{John E. Miller \\
  Artificial Intelligence/Engr. \\
  Pontificia Universidad Católica del Perú \\
  San Miguel, Lima, Peru \\
  \texttt{jemiller@pucp.edu.pe} \\\And
  Johann-Mattis List \\
  Chair of Multil. Comput. Linguistics / DLCE \\
  University of Passau / MPI-EVA \\
  Passau / Leipzig, Germany \\
  \texttt{mattis.list@uni-passau.de} \\}

\begin{document}
\maketitle
\begin{abstract}
% Language contact is a pervasive phenomenon prominently reflected in the borrowing of words from donor to recipient languages. So far, most computational approaches to borrowing detection treat all languages under study as equally important, even though dominant languages have a stronger impact on heritage languages than vice versa. Here we test new methods for lexical borrowing detection in contact situations where dominant languages play an important role. We apply two classical sequence comparison methods and one machine learning method to a sample of seven Latin American languages which have all borrowed extensively from Spanish. All systems perform well, with the supervised machine learning system modestly outperforming the classical systems.
% A review of detection errors shows that future methods could be substantially improved by increasing the data basis for dominant languages and by developing methods for borrowing detection which take donor words with divergent meanings into account.
Language contact is a pervasive phenomenon reflected in the borrowing of words from donor to recipient languages. 
Most computational approaches to borrowing detection treat all languages under study as equally important, even though dominant languages have a stronger impact on heritage languages than vice versa. 
We test new methods for lexical borrowing detection in contact situations where dominant languages play an important role, applying two classical sequence comparison methods and one machine learning method to a sample of seven Latin American languages which have all borrowed extensively from Spanish. All methods perform well, with the supervised machine learning system outperforming the classical systems.
A review of detection errors shows that borrowing detection could be substantially improved by taking into account donor words with divergent meanings from recipient words.
\end{abstract}

\section{Introduction}

Language contact is one of the most pervasive linguistic phenomena. 
It is the first factor that needs to be excluded when searching for genetic relations among the languages of the world, and it needs to be controlled for when searching for cross-linguistic universals. 
Any study trying to explain how humans use language must take language contact into account. It is also a unique witness of ancient contacts in human prehistory.
% Language contact is one of the most pervasive linguistic phenomena. Language contact is the first factor that needs to be 
% excluded when searching for genetic relations among the languages of the world. Language contact needs to be
% controlled for when searching for cross-linguistic universals. Finally, any study trying to explain how humans
% use language must take language contact into account. Due to its pervasiveness, language contact
% is also a unique witness of ancient contacts in human prehistory.

% Language contact is most prominently reflected in \emph{lexical borrowing}, the \emph{transfer} of words from one language (the donor language) to another (the recipient language). 
Language contact is most prominently reflected in \emph{lexical borrowing}, the \emph{transfer} of words from a donor language to a recipient language. Although research on the computational handling of lexical borrowing has made some progress of late, computational approaches to the investigation of language contact are still in their infancy \citep{List2019d}. Specifically methods that could infer the direction of borrowings have not been proposed so far. While numerous case studies investigate the influence of dominant languages on heritage languages \citep{Prochazka2017,Meisel2018}, we lack automated methods that could be used to study the influence of particular dominant languages in linguistically diverse areas of the world.   
 
% Given that the 
The number of standardized cross-linguistic wordlist collections has greatly increased over the past years \citep{Rzymski2020,List2022e}, with standard formats \citep{Forkel2018a} accompanied by software tools that allow scholars to prepare and curate standardized wordlists in an efficient manner \citep{Forkel2020}. 
%it would be particularly useful to have a method that allows us to detect which words in a cross-linguistic region have been borrowed from a dominant language.
It would be useful to have a computer assisted tool for linguists to detect which words in a cross-linguistic region have been borrowed from a dominant language and with further advances, a tool for inclusion in language contact assessment or in computational cladistics workflows. 
 
Here, we compare the suitability of three different systems to address this task -- two systems based on classical algorithms for automated sequence comparison that can be applied in supervised and unsupervised settings, and one supervised system based on extended machine learning techniques. We test these systems on a newly derived dataset of seven Latin American languages (Fig.~\ref{fig:map-1}) in which Spanish is a dominant donor language. Our results show that the supervised machine-learning system outperforms the classical systems.
%, but the differences are small.
%, and one of the classical systems also yields promising results.
% However, the differences are small, and one of the classical systems yields promising results.

\section{Previous Work}

Although still few in number, automatic methods for borrowing detection have been increasingly applied and developed in the past years. 
Early studies by \citet{Ark2007} and later \citet{Mennecier2016} compute edit distances between words from genetically unrelated languages and compare distances to thresholds,
in order to detect borrowed words in multilingual wordlists. % Borrowings are detected with the help of thresholds.

Besides edit distance, which directly calculates distances between phonetic sequences, sound class based methods cluster phonetic segments into sound classes and then compute distances between sound class sequences. The sound-class based alignment method (SCA)~\citep{List2012} provides sound class categories, scoring functions for distance measures, and modifiable gap scores based on prosodic context.

\citet{Zhang2021} compare edit distance performance against 
% distances derived from the Sound-Class-Based Alignment (SCA) method \citep{List2012}, 
SCA \citep{List2012} distance performance,
finding that SCA outperforms edit distance in accuracy. \citet{HantganA:2022} build on this work, using dedicated methods for automated cognate detection applied to languages from different language families in order to identify clusters of related words resulting from lexical borrowing.
\citet{List2022c} expand this work further, by applying a two-stage workflow in which they first identify language-family-internal cognates, using a method specifically apt for the detection of deep cognates, and then compute SCA distances between cognate sets from genetically unrelated languages in order to infer sets of words related by lexical transfer.
\citet{Miller2020} compute language models for inherited and borrowed words for individual languages from the World Loanword Database (WOLD, \citealt{WOLD}) using Markov Chains and Recursive Neural Networks and compare cross-entropies for inherited and borrowed language models in order to identify borrowings from monolingual information alone.
%In monolingual WOLD \citep{WOLD_zenodo:2019} wordlists, \citep{Miller2020} trained Markov chain and Recursive Neural Network (RNN) language models, self-supervised, to estimate word cross-entropies with inherited and borrowed word models competed to assign borrowed status. 
%Segmented IPA training wordlists were split into inherited and borrowed words and resulting inherited and borrowed word language models detected borrowings from held-out test datasets. 
%The approach was monolingual, training is minimally supervised, and inference is to test datasets in a 10-fold cross-validation.  The slightly better RNN model attains a cross-validation $\textrm{F1 score}=0.604$. 

% In monolingual WOLD \citep{WOLD_zenodo:2019} wordlists, \citep{Miller2020} trained Markov chain and Recursive Neural Network (RNN) language models, self-supervised, to estimate word cross-entropies. Segmented IPA training wordlists were split into inherited and borrowed words and resulting inherited and borrowed word language models detected borrowings from held-out test datasets. The approach was monolingual and minimally supervised with the slightly better RNN model attaining a cross-validation $\textrm{F1 score}=0.604$. 

\citet{KaipingG:2022b} use automated methods for cognate detection \citep{List2017c} on a target set of Timor-Alor-Pantar languages. In order to infer borrowings from Indonesian and Tetun (not in the target set), they include both languages in their sample and treat all cognate sets that involves words from either of the two languages as borrowings. \citet{Moro2023} apply a similar approach to investigate borrowings in Alorese.

\citet{MiC:2020} and \citet{NathA:2022} train binary classifiers, mainly neural based, on large wordlists to predict borrowed words, and achieve F1 scores in the 0.75 to 0.85 range.
Their workflows seem cumbersome, compute intensive, and not minimalist, but the results are promising.  

\section{Materials and Methods}
\subsection{Materials}
\begin{figure}[tb]
\centering
\includegraphics[width=\linewidth]{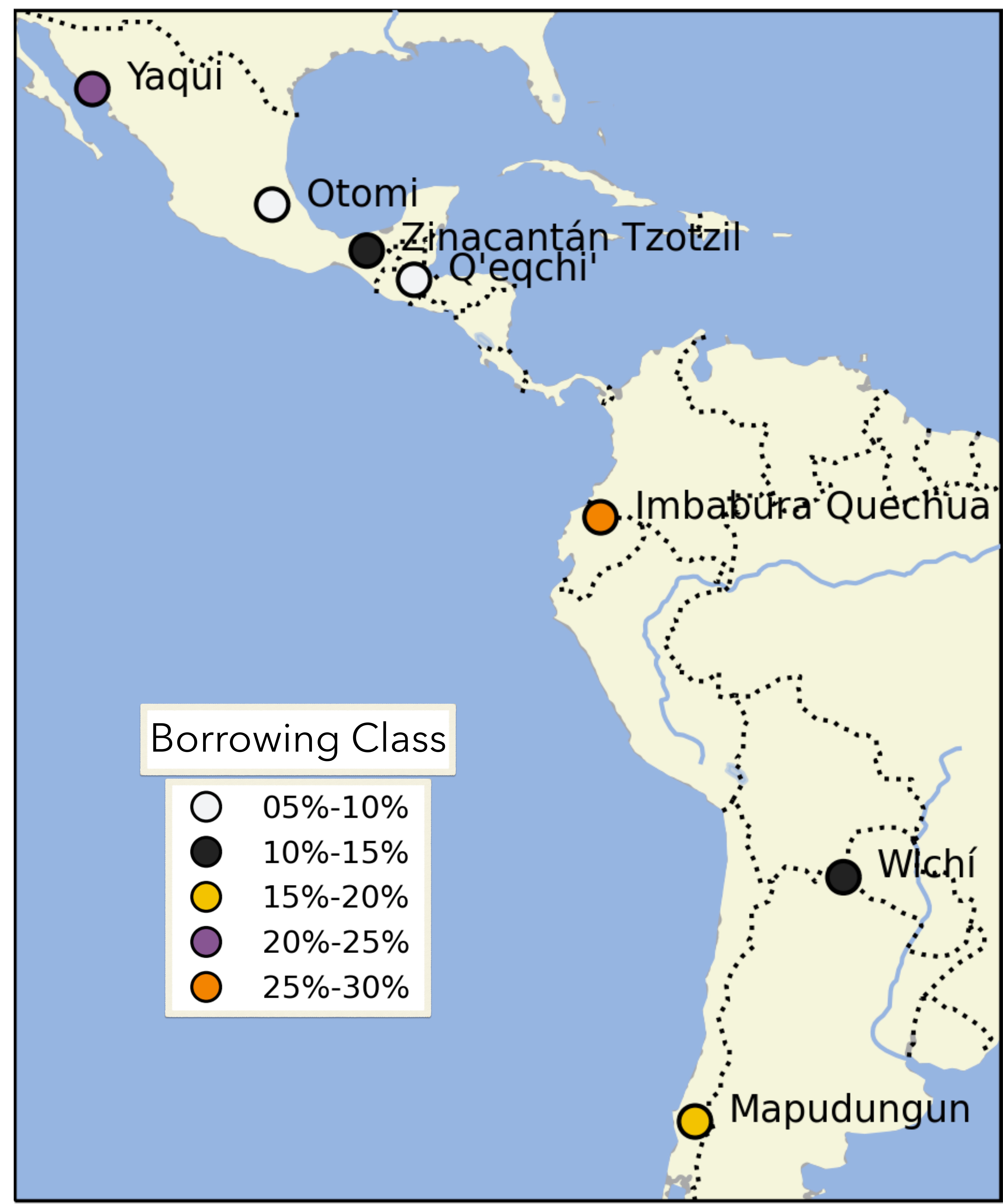}
\caption{Map of languages with Spanish borrowing class.}
\label{fig:map-1}
\vspace{-1em}
\end{figure}

For this study, a new comparative wordlist was created by taking data for seven Latin American languages from WOLD (\url{https://wold.clld.org}, \citealt{WOLD}) and combining them with a wordlist of Spanish derived from the Intercontinental Dictionary Series~(\url{https://ids.clld.org}, \citealt{IDS:2015}).
% While phonetic transcriptions for the Latin American language varieties had been created in previous work \citep{Miller2020}, 
Phonetic transcriptions for the Latin American languages were added to WOLD by \citet{Miller2020}. 
Latin American Spanish phonetic transcriptions were added for this study (and could be later expanded by adding more transcriptions from historical varieties of Spanish). 
The resulting dataset conforms to the standards suggested by the Cross-Linguistic Data Formats initiative (CLDF, \url{https://cldf.clld.org},  \citealt{Forkel2018a}). The data curation follows the Lexibank workflow \citep{List2022e} and checks that data conform to certain standards, with languages being linked to Glottolog (\url{https://glottolog.org}, \citealt{Glottolog}, Version 4.7), concepts being linked to Concepticon (\url{https://concepticon.clld.org}, \citealt{Concepticon}, Version 3.0), and transcriptions following the B(road)IPA conventions of the Cross-Linguistic Transcription Systems reference catalog (\url{https://clts.clld.org}, \citealt{CLTS}, Version 2.2, see \citealt{Anderson2018}). Details of the resulting database are shown in the map of language locations along with percentages for borrowings from Spanish in Fig. \ref{fig:map-1} and in Tab. \ref{tab:databasedetails}. Q'eqchi' and Zincantán Tzotzil are both Mayan languages, but appear substantially varied in the database.

%The Spanish largely orthographic coding was converted to segmented IPA for consistency with WOLD via an orthographic profile~\citep{MoranS:2018} that we developed.
% - WOLD - 7 Latin American languages with major Spanish Influence
% - IDS - Spanish language table, forms upgraded to segmented IPA
% - CLDF dataset, SABor, constructed from relevant WOLD and IDS languages.

% \begin{figure}[tb]
% \centering
% \includegraphics[width=\linewidth]{map-1.png}
% \caption{Location of languages in our dataset.}
% \label{fig:map-1}
% \vspace{-1em}
% \end{figure}

% \begin{table}[!ht]
%     \centering
%     \begin{tabular}{lrrr}
%     \hline
%         Property & Overall & Min & Max \\ \hline
%         Concepts & 1,308 & 955 & 1,308 \\
%         Lexemes & 12,100 & 1,156 & 1,773 \\
%         IPA Tokens & 72,550 & 7,129 & 11,730 \\
%         IPA Vocabulary & 112 & 28 & 57 \\
%     \end{tabular}
%     \caption{Database details for seven Latin American languages plus Latin American Spanish.}
%     \label{tab:databasedetails}
% \end{table}

\begin{table}[!ht]
    \setlength{\tabcolsep}{4pt}
    \centering
    \resizebox{\linewidth}{!}{
    \begin{tabular}{l @{\hspace{0.2\tabcolsep}} r
    @{\hspace{0.9\tabcolsep}} r
    @{\hspace{0.9\tabcolsep}} r
    @{\hspace{0.9\tabcolsep}} r}
    \hline
        \textbf{Language} & \textbf{Concepts} & \textbf{Lexemes} & \textbf{Segments} & \textbf{Vocab.} \\ \hline
        Imb. Quechua & 1,155 & 1,156 & 7,177 & 33 \\
        Mapudungun & 1,040 & 1,242 & 7,356 & 33 \\
        Otomi & 1,252 & 2,241 & 11,730 & 57 \\
        Q'eqchi' & 1,211 & 1,773 & 10,367 & 49 \\
        Wichí & 1,128 & 1,219 & 8,233 & 44 \\
        Yaqui & 1,242 & 1,433 & 9,297 & 28 \\
        Zin. Tzotzil & 955 & 1,266 & 7,129 & 41 \\
        Spanish & 1,308 & 1,770 & 11,261 & 30 \\ \hline
        Aggregate & 1,308 & 12,100 & 72,550 & 112 \\ %\hline
    \end{tabular}}
    \caption{Database details for seven Latin American languages plus Latin American Spanish.}
    \label{tab:databasedetails}
    \vspace{-1em}
\end{table}

\subsection{Methods}
\paragraph{Methods for Borrowing Detection.}
We develop three different methods for the detection of borrowings \emph{from} a dominant language \emph{to} non-dominant languages in multilingual wordlists. Following historical linguistics comparative method practice~\citep{CampbellL:2013}, only word forms corresponding to the same concept are considered as candidates for borrowing.

The first method, called \emph{Closest Match} borrowing detection in the following, iterates over all word pairs that express the same concept in the dominant language and the heritage languages and then computes phonetic distances. Word pairs whose phonetic distance is below a certain threshold are judged to be borrowings from the dominant language. We test two phonetic distances, the normalized edit distance (NED) -- the classical edit distance \citep{Levenshtein1965} between two words, divided by the length of the longer word -- and the SCA distance \citep{List2012}.

The second method, called \emph{Cognate-Based} borrowing detection in the following, follows the approach by \citet{HantganA:2022}: it first computes cognates using a cluster-based approach for automated cognate detection in which words expressing the same concept whose average phonetic distance is below a certain threshold are assigned to the same cognate set \citep{List2017c}, and then identifies all words assigned to cognate sets involving the dominant language as borrowings. We tested again normalized edit and SCA distances.
% two phonetic distances, normalized edit distance and SCA distance. 

The third method, called \emph{Classifier-Based} borrowing detection in the following, iterates over all word pairs with the same concept, but stores phonetic distance scores for various distance measures as vectors, which can then be used to train a classifier, firstly, a linear Support Vector Machine (SVM) \citep{CristianiniN:2000}, in a supervised setting. 
We tested various phonetic distance measures, but report only on the combination of normalized edit and SCA distances, as they yielded the best results.
Both Closest Match and Cognate-Based methods require a fixed threshold which we estimate from the training data. So all three methods are considered as supervised.

\paragraph{Sampling.}
Train-test splits are made based on concepts rather than individual word entries.  This permits matching of words for the same concept in % both pairwise and cognate based 
in all methods without loss of candidate words. Treating train-test split as a nuisance variable takes into account differences between partitions across methods thus controlling for effects of sampling by concepts with differing borrowing behavior or statistical dependencies between test partitions due to sampling without replacement. See~\citep{drorR:2018} for mention of the dependency problem with cross-validation. Our use of a fixed partition across treatments and analysis of variance controlling for partition as a nuisance or `blocking' variable accounts for this dependency, and takes advantage of any systematic effects in borrowing behavior by partition.

\paragraph{Evaluation.} For the analysis of the cross-validation data, we use a randomized blocks design where experiment is the treatment or factor, and test partition is the randomized block or nuisance variable. A standard analysis of variance partitions treatment effects, nuisance variable, and error, and permits a more powerful test of treatment differences without the nuisance variable variance. We follow up statistically significant findings for treatment (experiment), with comparisons of experiments versus the overall average using a joint (family) error rate.
\paragraph{Implementation.}
Our methods are implemented in Python, making specifically use of the CLDFBench package (\url{https://pypi.org/project/cldfbench/}, \citealt{Forkel2020}, Version 1.13.0) to provide commandline access to all methods described here. For the computation of alignments and edit distances, LingPy (\url{https://pypi.org/project/lingpy}, \citealt{LingPy}, Version 2.6.9) is used. SVM and evaluation are realized with the help of Scikit-Learn (\url{https://pypi.org/project/scikit-learn/}, \citealt{scikit-learn:2011}, Version 1.2.1).

\section{Results and Discussion}
We tested our three methods with two distance measures in five experiments 
(normalized edit and SCA distances individually in both Closest Match and Cognate-Based methods, and combined in the Classifier-Based method) using a 10-fold cross-validation on our data and reporting precision, recall, F1 scores, accuracy, and execution times (mm:ss). F1 score is the primary result measure; accuracy and execution time are informational. 
%~\footnote{Average Spanish borrowing of 15.2\% gives a majority decision accuracy of 84.8\% as an accuracy baseline.}
%\footnote{The cross-validation uses 10 fixed train/test splits where in each split 1/10\textsuperscript{th} of the data is in test and does not overlap with any other test split. 
The 10-fold cross-validation uses the same 10 fixed train and non-overlapping test splits for all experiments. With few parameter estimates (1 threshold each for Closest Match and Cognate-Based, 2 distance and 7 target language coefficients for Classifier), a separate \textit{train} split into \textit{fit/val} is not necessary. 

All methods perform well with less than 5 points separating the highest from the lowest F1 scores. 
Tab.~\ref{tab:10-fold-CV} shows the results of the ten-fold cross validation of our three methods in five experiments. An analysis of variance,\footnote{Statistical analyses with JMP \citep{JMP:2021}.} 
with experiment as the effects variable and train-test split as the nuisance variable, shows highly significant effects for precision ($\textrm{F}_{4,36}=25.74, p<0.0001$) and F1 score ($\textrm{F}_{4,36}=14.3, p<0.0001$).
% \footnote{F1 dependency on precision is reflected in the test result.} but not for recall. 

% As can be seen from the table, 
Closest Match with normalized edit distance performs poorly, while Classifier-Based with combined normalized edit and SCA distances performs well. Classifier-Based performs better than the average of all experiments in F1 score, and substantially better in precision versus other experiments; the method is conservative, with a low number of false positives. Performance on remaining experiments is indistinguishable from the overall average of all experiments combined. The Cognate-Based method is compute intensive performing multiple alignment over all languages. Accuracy is well above the majority decision accuracy of 84.8\% ($100\%-15.2\%\textrm{ borrowing}$) in all experiments.

\begin{table}[!ht]
    \setlength{\tabcolsep}{4pt}
    \centering
    \begin{tabular}{l @{\hspace{0.5\tabcolsep}} rrrrr}
    \hline
        \textbf{Method} & \textbf{Prec.} & \textbf{Rec.} & \textbf{F1} & \textbf{Acc.} & \textbf{mm:ss}\\ 
        \hline
        \multicolumn{2}{l}{\textbf{Closest Match}} \\
        NED  & \underline{0.832} & 0.703 & \underline{0.761} & 0.938 & 00:15\\
        SCA & 0.869 & 0.720 & 0.787 & 0.945 & 00:29\\
        \multicolumn{2}{l}{\textbf{Cognate-Based}} \\
        NED & 0.853 & 0.705 & 0.771 & 0.941 & 01:48\\
        SCA & 0.862 & 0.719 & 0.783 & 0.944 & 04:49\\
        % \multicolumn{2}{l}{\textbf{Classifier-Based}} \\
        \multicolumn{4}{l}{\textbf{Classifier-Based} SVM (linear)} \\
        NED, SCA & \textbf{0.931} & 0.713 & \textbf{0.806} & 0.952 & 00:37\\
        % \hline
        \hline
    \end{tabular}
    \caption{Ten-fold cross-validation for three methods with NED (normalized edit) and SCA (Sound-Class based phonetic alignment) distance measures. \textbf{Bolded} estimates are superior to and \underline{underlined} estimates inferior to the the overall average using analysis of means~\citep{NelsonP:2005} with joint error rate $\alpha=0.05$.}
    \label{tab:10-fold-CV}
    \vspace{-1em}
\end{table}

A search for classifier improvements
%\footnote{Thanks to reviewer suggestions.} 
prompted several \textit{ad hoc} experiments (see tab.~\ref{tab:10-fold-CV-experiments}). We observe:
(1)~A radial basis function (rbf) SVM classifier performs no better than our linear SVM. We suspect the estimated target language parameters do not generalize well to held-out data.
(2)~A logistic regression classifier performs on par with our linear SVM.
(3)~A weight balanced SVM classifier trades an increase in recall for a larger drop in precision.  
%Overall F1 score performance suffers.
We also test whether using separate trials for each target language in Closest Match, would perform as well as all languages together. A combined trial performs better; a single threshold estimate appears to generalize better to held-out data than using individual language estimates.

\begin{table}[!ht]
    \setlength{\tabcolsep}{4pt}
    \centering
    \begin{tabular}{l @{\hspace{0.3\tabcolsep}} rrrr}
    \hline
        \textbf{Experiment} & \textbf{Prec.} & \textbf{Rec.} & \textbf{F1} & \textbf{Acc.} \\ \hline
        \multicolumn{5}{l}{\textbf{Classifier Variations - NED, SCA}} \\ 
        SVM (rbf) & 0.945 & 0.694 & 0.799 & 0.951 \\ 
        Logistic regression & 0.914 & 0.728 & 0.809 & 0.952 \\ 
        SVM (balanced) & 0.613 & 0.826 & 0.704 & 0.902 \\ 
        \multicolumn{5}{l}{\textbf{Closest Match - SCA}} \\ 
        Each language (avg) & 0.860 & 0.707 & 0.770 & 0.941 \\ 
        % SVM Classifier - NED,SCA & 0.936 & 0.639 & 0.790 & 0.949 & 4:29 \\ 
        \hline
    \end{tabular}
        \caption{Ten-fold cross-validation for several \textit{ad hoc} experiments  with NED (normalized edit) and SCA (Sound-Class based phonetic alignment) distance measures. Classifier experiments: SVM with radial basis function, Logistic regression, linear SVM with balanced class weights. Descriptive statistics only.} 
        %SVM classifier with radial basis function, Logistic regression classifier, linear SVM classifier with balanced class weights. Descriptive statistics only.
    \label{tab:10-fold-CV-experiments}
    \vspace{-1em}
\end{table}

%\subsection{Results by Target Language}
Tab.~\ref{tab:language-detail} shows the results of the Classifier-Based method for the seven target languages in our sample with training and evaluation over the entire dataset. 
%As can be seen from the table,
There is some variation in performance by language, in particular, with recall in $[0.615, 0.778]$. We detect %(\textit{ad hoc} test) 
a linear relation between the performance and the amount of borrowings from the dominant language in the target languages. (Precision: $\textrm{r}=-0.39, \textrm{NS}$; Recall: $\textrm{r}=0.88, p<0.01$; F1 score: $\textrm{r}=0.85, p<.01$; 1-sided Pearson correlation tests with $df=5$). Recall and F1 scores improve as borrowing increases. This could be an artifact of higher borrowing resulting in better estimation of a target language coefficient, or more interestingly, a cultural process where more dominant-donor borrowing corresponds to reduced phonetic adaption into the target language.

%It doesn't appear due to improved estimation with more borrowing, since there are so few parameters to estimate.  Perhaps, there is a cultural explanation. More borrowing could result in less adaption to the target language, with subsequent easier recall of borrowed words.

\begin{table}[!ht]
    \setlength{\tabcolsep}{4pt}
    \centering
    \resizebox{\linewidth}{!}{
    \begin{tabular}{l @{\hspace{0.2\tabcolsep}} r
    @{\hspace{0.9\tabcolsep}} r
    @{\hspace{0.9\tabcolsep}} r
    @{\hspace{0.9\tabcolsep}} r
    @{\hspace{0.9\tabcolsep}} r}
    \hline
        \textbf{Language}   & \textbf{Prec.} & \textbf{Rec.} & \textbf{F1} &  \textbf{Acc.} & \textbf{Borr.} \\ 
        \hline
        % Imbabura Quechua    
        Imb. Quechua      & 0.921 & 0.773 & 0.841 & 0.924 & 26\% \\
        Mapudungun        & 0.944  & 0.716 & 0.814 & 0.950 & 15\% \\
        Otomi               & 0.932 & 0.692 & 0.794 & 0.968 & 9\% \\
        Q'eqchi'              & 0.934 & 0.615 & 0.742 & 0.961 & 9\% \\
        Wichí               & 0.952 & 0.658 & 0.778 & 0.953 & 12\% \\
        Yaqui               & 0.938 & 0.778 & 0.851 & 0.941 & 22\% \\
        % Zinacantan Tzotzil         
        Zin. Tzotzil        & 0.932 & 0.661 & 0.773 & 0.949 & 13\% \\ \hline
        Average             & 0.934 & 0.714 & 0.810 & 0.952 & 15\% \\ % \hline
    \end{tabular}}
    \caption{Individual language results for the Classifier-Based borrowing detection methods on the seven target languages in our sample. The last column shows the proportion of Spanish borrowings.}
    \label{tab:language-detail}
    \vspace{-1em}
\end{table}

\begin{figure}[!bt]
\centering
\includegraphics[width=0.95\linewidth]{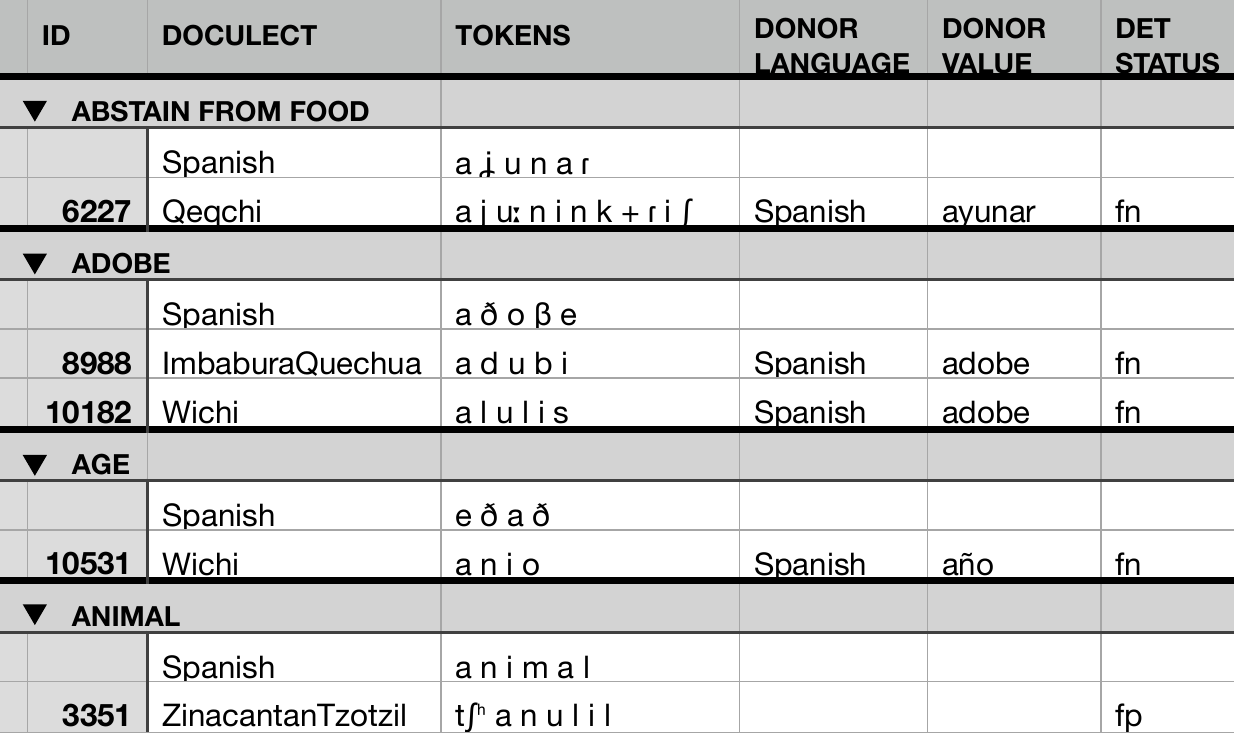}
\caption{Example collection of detection errors.}
\label{fig:detail-evaluate-example}
\vspace{-1em}
\end{figure}
%\paragraph{Error Analysis} 
To get a better understanding about the different types of errors that our best performing experimental combination commits, we conducted a detailed error analysis from the Classifier-Based borrowing detection results.
%with training (analysis) over the entire dataset. 
A spreadsheet snippet (Fig. \ref{fig:detail-evaluate-example}),  %showing evaluation output, % how we reviewed evaluation output for errors, 
serves as a reference for several error types.
For undetected borrowings (false negatives), we identified four error types:
(1)~cases where the borrowed form was not present in the donor wordlist, e.g., Mapudungun \textit{\textipa{peso}} ``coin'' is borrowed from Spanish \textit{peso} ``peso'', but our Spanish wordlist only has \textit{moneda}, 
(2)~cases where the form was present in the donor wordlist, but with a different concept,
%(and therefore undetected by our methods that compared words within concept slots), 
e.g., Wichi \textit{\textipa{anio}} ``age'' is borrowed from Spanish \textit{año} ``year'', while the Spanish word for ``age'' is \textit{edad}, 
(3)~cases of large phonetic distance between donor and recipient forms, e.g., Wichi \textit{alulis} ``adobe'', which is somewhat distant from Spanish \textit{adobe}, and
(4)~cases of unrecognized partial borrowing, e.g., Qeqchi \textit{\textipa{aiunink-riS}} ``abstain from food'', which is partially borrowed from Spanish \textit{ajunar} ``fast''.
For falsely detected borrowings (false positives), we identified three error types:
(1)~cases where the form was not borrowed from the dominant language but \emph{vice versa}, e.g., Spanish \textit{poroto} ``bean'' was borrowed \emph{from} Quechua \textit{\textipa{purutu}},
(2)~cases of chance similarities between word forms, e.g., Spanish \textit{animal} ``animal'' and Zinacantan Tzotzil \textit{\textipa{tSanulil}}, and 
(3)~cases so improbably similar that we suspect errors in the original annotation, e.g., Spanish \textit{pelota} ``ball'' and Wichi \textit{pelutaj}.

For a large sample of concepts, we tallied 139 undetected (false negative) and 26 falsely detected (false positive) borrowings (see Tab. \ref{tab:error-diagnostics}).
Most errors were in recall, with many of these borrowings from lexemes \textbf{not} within the same concept.

%We tallied detection errors by concept, in a sample of 139 undetected (false negative) and 26 falsely detected (false positive) borrowings (see Tab. \ref{tab:error-diagnostics}).
%The vast majority of errors were in recall, with many of these due to borrowings from lexemes \textbf{not} within the same concept.

% semantic shift or lack of representation in wordlists.

% In a systematic sample of 115 concepts with errors from the Classifier results, We tallied detection errors, with 139 undetected (false negative) and 26 falsely detected (false positive) borrowings (Tab. \ref{tab:error-diagnostics}). 
% For undetected borrowings: 
% (1)~half were due to the donor form coming from a different concept than the recipient form (75),  
% (2)~a fifth were due to the donor wordlist lacking the borrowed form (28), and
% (3)~a fifth were due to the large phonetic distance between donor and borrowed forms (31). 

% For the few falsely detected borrowings, three quarters were due to the unexpected similarity of forms, by chance (10) or possible data error (9). Importantly, only 7 errors overall were because the \emph{borrowing was not} from the dominant language. 

\begin{table}[!tb]
    \setlength{\tabcolsep}{4pt}
    \centering
    \begin{tabular}{@{\hspace{1.0\tabcolsep}} l
    @{\hspace{-1.0\tabcolsep}} rr}

    \multicolumn{2}{c}{\textbf{Undetected Borrowings}} \\\hline 
    % -- False Negatives}} \\
        Error Type & Count & Pct\\ \hline
        borrowed form not in donor list  & 28 & 17\\
        % different - related concept & 67 \\
        different concept than recipient form & 75 & 45\\
        % different - independent concept & 8 \\
        large phonetic distance & 31 & 19\\
        partial borrowing as only reason & 5 & 3\\
        \textbf{Subtotal} & 139 & 84\\ \hline
    \multicolumn{2}{c}{\textbf{Falsely Detected as Borrowings}} \\\hline 
        %-- False Positives}} \\
        Error Type & Count & Pct\\ \hline
        % not borrowed from dominant donor & 7 \\
        direction not from dominant donor & 7 & 4\\
        chance similarity of form & 10 & 6\\
        likely dataset error & 9 & 5\\
         \textbf{Subtotal} & 26 & 16\\ \hline
        \textbf{Total} & 165 & 100\\ % \hline
        % partial borrowing - additional factor & 13 \\ \hline
    \end{tabular}
    \caption{Summary over sample of undetected (false negative) and falsely detected (false positive) borrowings.}
    \label{tab:error-diagnostics}
    \vspace{-1em}
\end{table}
% 1. Individual experiments to determine optimal threshold and method configurations. For discussion. Summary table and chart of effects.
% 2. 10-fold cross validations of major contenders for each method. Summary tables.
% 3. Results by individual target languages for best model on all training data. Summary table.
% 3. Detail test results for fold 0 of test partition for purpose of diagnosing where errors occurred and possible improvements.

% \section{Discussion}
% * For now just a collection discussion points. *
% Results by language showed a linear relationship between proportion of borrowing and detection performance most notably in recall. Why is this?  It doesn't appear due to improved estimation with more borrowing, since there are so few parameters to estimate.  Perhaps, there is a cultural explanation. More borrowing could result in less adaption to the target language, with subsequent easier recall of borrowed words.

\section{Conclusion}
How well can we automatically detect borrowings from dominant languages based on wordlist data? 
We devised three general methods to detect borrowed words from dominant languages, 
two based on sequence comparison workflows 
%(Closest Match and Cognate-Based) 
and one based on a classifier. %(Classifier-Based).
%two based on direct workflows involving sequence comparison (Closest Match and Cognate-Based borrowing detection) and one based on a classifier (Classifier-Based borrowing detection).
The classifier-based method showed the best performance, with F1 scores of 0.81, and high precision of 0.93.
This method could already prove very useful in computer-assisted workflows. 
% Given that methods for the detection of borrowings from dominant languages already play an important role in studies on phylogenetic reconstruction \citep{Kaiping2022b}, with dominant languages having an important impact on the evolutionary dynamics of language history in many places of the world, our results provide important insights for future work in this direction. 
Our investigation of detection errors shows several opportunities for improvement. 
Most undetected borrowings result from current methods' restriction to searching only for word forms for the same concept.
% restricting the search for borrowings to word forms expressing identical concepts in our wordlists. 
We see great potential in improvements that account for borrowing accompanied by \emph{semantic shift}, specifically: 
% Designing such methods is not trivial, however, since an unconstrained comparison of word forms regardless of their meaning would also increase the number of false positives.
%Additional future improvements include (1)~the use of refined phonetic distances (both for the two methods based on sequence comparison and the classifier-based approach), and (2)~the design of methods that search for partial similarities between sequences.
% Potential improvements include 
%(1)~augment donor wordlist coverage of possible forms, (2)~relax ``same concept'' to a ``similar concept'' requirement for matching forms, or (3)~ fit language models to wordlists and incorporate word cross-entropy into the classifier without the ``same concept'' restriction. 
(1)~augment donor wordlist coverage of possible forms, (2)~relax ``same'' to  ``similar''  concept restriction for matching forms, or (3)~ fit language models to wordlists and add word cross-entropy to the classifier without restriction. 
Tests adding word cross-entropy, not reported here, look very promising, but more research is needed. 
%to check them thoroughly.

%(3)~fit language models to wordlists and incorporate word cross-entropy into the classifier without a same concept restriction.\footnote{Recent results incorporating word cross-entropies into the classifier along with NED, SCA, show substantial improvement in borrowing detection (precision=0.871, recall=0.827, F1 score=0.848, accuracy=0.958) in 10-fold cross-validation.}
%monolingual borrowing detection into the classifier.

%\textit{most errors of recall} (false negatives) result from not detecting a borrowing from a related concept or not having defined all donor concept forms.  
%These are errors that a competent linguist would not make, but an algorithmic procedure needs to have these data accessible for use in borrowing detection. %This will guide our future work.

%We demonstrated competitive methods for detecting lexical borrowings from a dominant donor, including by implication the direction of borrowing (from the donor). We have shown that the a classifier combining multiple distance methods performs better than stand alone closest match and cognate based methods. Importantly we have shown the potential of three avenues for substantial improvement: 
%\begin{enumerate*}
%\item incorporate more varied functions for measuring distance into the classifier, 
%\item incorporate related concepts into tests of borrowing, and
%\item make donor wordlist representation of forms more exhaustive.
%\end{enumerate*}

\section*{Limitations}
%Methods used here for detection of lexical borrowing, can be applied more generally, to language contact and borrowing. 
In this study, we apply limited methods for detection of lexical borrowing to the case of a single dominant donor language (Spanish) wordlist 
%(Spanish using Latin American pronunciation) 
versus seven Latin American language wordlists. 
This application uses relatively sparse data, with an average of 1,512 lexemes per language wordlist, with very few estimated parameters, and so apt for lower resource languages. Future work will seek to remove the wordlist limitation.

\section*{Ethics Statement}
Our data are taken from publicly available sources. There are no ethical issues or conflicts of interest in this work.  

\section*{Supplementary Material}
The supplementary material accompanying this study contains the data and code needed to replicate the results reported here, along with detailed information on installing and using the software. It is curated on GitHub (\url{https://github.com/lexibank/sabor}, Version 1.0) and has been archived with Zenodo (\url{https://doi.org/10.5281/zenodo.7591335}).

\section*{Acknowledgements}
This research was supported by the Max Planck Society Research Grant \textit{CALC³} (JML, \url{https://digling.org/calc/}), the ERC Consolidator Grant \textit{ProduSemy} (JML, Grant No. 101044282, see \url{ https://cordis.europa.eu/project/id/101044282}), and by the Graduate School of the Pontificia Universidad Católica del Perú (PUCP) through the Huiracocha-2019 scholarship program, see \url{https://posgrado.pucp.edu.pe} (JEM). Views and opinions expressed are however those of the author(s) only and do not necessarily reflect those of the European Union or the European Research Council Executive Agency (nor any other funding agencies involved). Neither the European Union nor the granting authority can be held responsible for them.
We thank the anonymous reviewers for helpful comments and all people who share their data openly, so we can use it in our research.

\bibliography{anthology,custom}
\bibliographystyle{acl_natbib}

% \appendix

% \section{Example Appendix}
% \label{sec:appendix}

% This is a section in the appendix.

\end{document}